\UseRawInputEncoding
\documentclass{article}

\usepackage[english]{babel}

\usepackage[letterpaper,top=2cm,bottom=2cm,left=3cm,right=3cm,marginparwidth=1.75cm]{geometry}

\usepackage{amsmath}
\usepackage{graphicx}
\usepackage[colorlinks=true, allcolors=blue]{hyperref}

\title{Language Cognition and Language Computation – Human and Machine Language Understanding\footnote{This paper is originally written in Chinese and published in SCIENTIA SINICA Informationis. Here we translate it into English with an extension of recent work.}}

\author{\small Shaonan Wang$^{1,2,*}$, Nai Ding$^{3,4,*}$, Nan Lin$^{5,6}$, Jiajun Zhang$^{1,2}$, Chengqing Zong$^{1,2}$}
\date{\small%
    $^1$National Laboratory of Pattern Recognition, Institute of Automation, CAS, Beijing, China\\%
    $^2$School of Artificial Intelligence, University of Chinese Academy of Sciences, Beijing, China\\%
    $^3$Key Laboratory for Biomedical Engineering of Ministry of Education, College of Biomedical Engineering and Instrument Sciences, Zhejiang University, Hangzhou, China\\%
    $^4$Zhejiang Lab, Zhejiang University, Hangzhou, China\\%
    $^5$CAS Key Laboratory of Behavioural Sciences, Institute of Psychology, Beijing, China\\%
    $^6$Department of Psychology, University of Chinese Academy of Sciences, Beijing, China\\%
    *Corresponding authors. Email: shaonan.wang@nlpr.ia.ac.cn, ding nai@zju.edu.cn
}

\begin{document}
\maketitle
\begin{abstract}
Language understanding is a key scientific issue in the fields of cognitive and computer science. However, the two disciplines differ substantially in the specific research questions. Cognitive science focuses on analyzing the specific mechanism of the brain and investigating the brain's response to language; few studies have examined the brain's language system as a whole. By contrast, computer scientists focus on the efficiency of practical applications when choosing research questions but may ignore the most essential laws of language. Given these differences, can a combination of the disciplines offer new insights for building intelligent language models and studying language cognitive mechanisms? In the following text, we first review the research questions, history, and methods of language understanding in cognitive and computer science, focusing on the current progress and challenges. We then compare and contrast the research of language understanding in cognitive and computer sciences. Finally, we review existing work that combines insights from language cognition and language computation and offer prospects for future development trends.
\end{abstract}

\section{Introduction}
Language is a multilevel symbolic system that includes multiple levels: phonetics, morphology, syntax, semantics, and pragmatics. The most basic language symbols can be combined to form more complex and endless symbol sequences to allow flexible expression of meaning. As such, language is also considered the carrier of human thought and the most natural tool through which humans exchange ideas and express emotions.

Because of the diverse and flexible characteristics of language, it is difficult to study the mechanism of human language understanding and to build a computation model that can understand language. In the early days of computer science, language research pioneers attempted to conduct cross-disciplinary research in computer science, linguistics, and cognitive science. They aimed to establish connections between human language-understanding mechanisms and language-computation models \cite{hebb2005organization,hinton1984distributed,hopfield1982neural,mcculloch1943logical,turing1950computing,chomsky1956three}. However, owing to the complexity of the problem, interdisciplinary research has gradually become separated over the decades, forming subfields such as natural language understanding in computer science, psycholinguistics in cognitive psychology, and neurobiology of language research in cognitive neuroscience. In this paper, "cognitive science" mainly refers to the two fields of cognitive psychology and cognitive neuroscience, particularly the branches of psycholinguistics and the cognitive neuroscience of language \cite{csaba2015kemmerer}.

  \begin{figure}[htbp] \centering
  \includegraphics[width=0.9\textwidth]{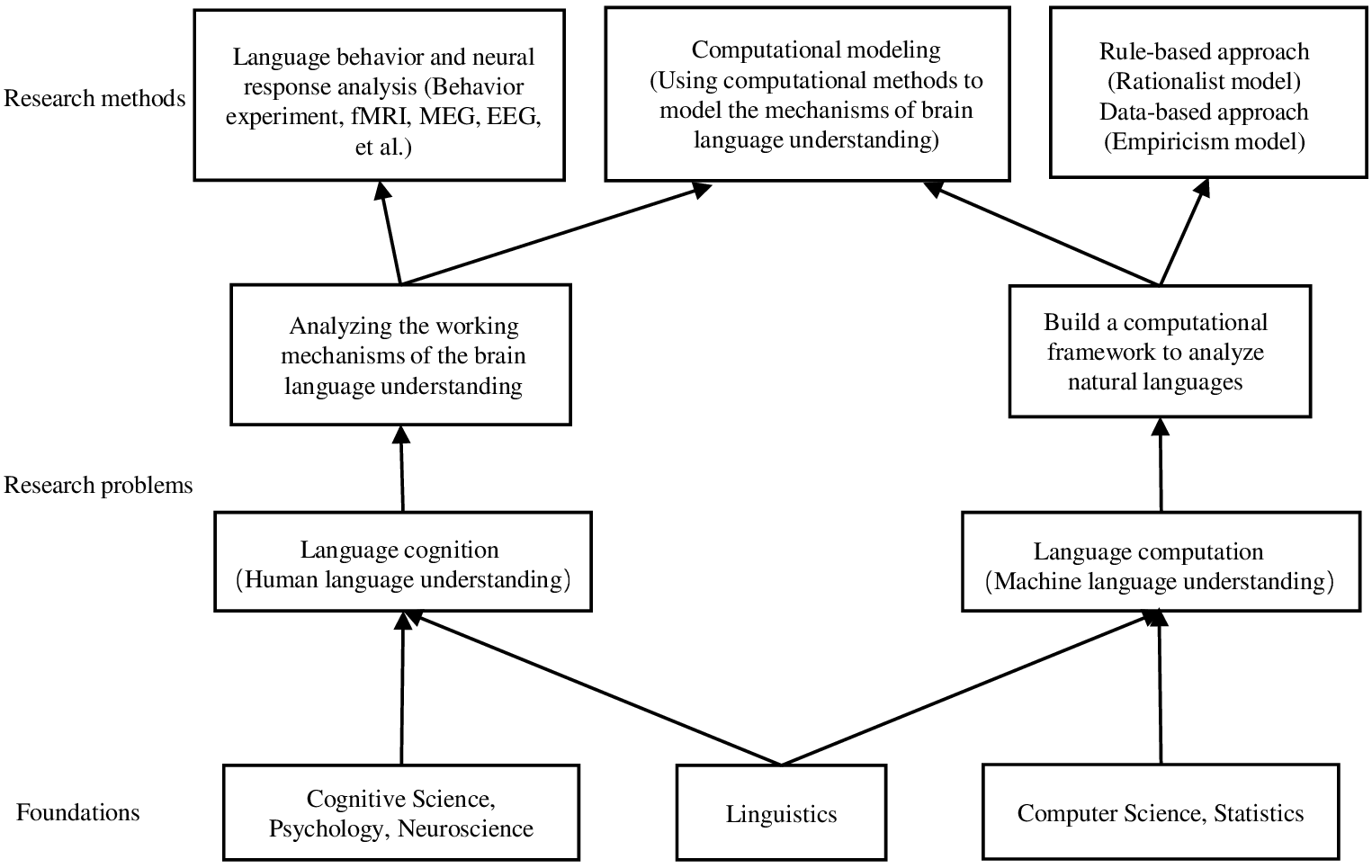}\\
  \caption{Connections between cognitive sciences and computer sciences on the language understanding problem}\label{fig1}
  \end{figure}
  
Figure \ref{fig1} shows the relationship between cognitive and computer science in the direction of language understanding. There are substantial differences in the research questions and methods adopted in the two fields. Computer scientists primarily adopt rationalist (represented by rule-based methods) and empirical methods (data-driven modeling methods, represented by statistical machine learning and neural network methods) and pay more attention to applied research--that is, how to build an intelligent system to understand natural language to complete various practical applications (such as machine translation, dialogue system, and automatic summarization). The "understanding" here refers, more precisely, to application-oriented "processing"; thus, in many cases, "natural language processing" is commonly used to collectively refer to this research direction. 

On the other hand, cognitive scientists adopt neuroimaging and behavioral analysis methods and pay more attention to the psychological and neural basis of human language understanding, such as the functions of each brain region in language understanding and how neural activities encode different levels of language information. The commonality between these two fields is that both use linguistics as the subject basis and computational modeling as tools for analysis in terms of language understanding. In general, language cognition and language computation have achieved fruitful results in their respective directions, and new theories and methods have been continuously proposed and successfully applied.

In recent years, improvements in computing resources and deep learning algorithms have led to the rapid development of artificial intelligence and computer science. Computers have defeated professional human players in tasks such as chess, quizzes, Go, and video games. In the field of natural language understanding, automatic dialogue systems and question answering systems, such as Siri and Watson, have also emerged as well as more practical machine translation systems \cite{hoy2018alexa,strickland2019ibm,zhu2019ncls}. However, the current artificial intelligence system relies on largescale training data that lack basic common-sense knowledge; thus, a large gap remains between human and artificial intelligence in terms of learning and generalization abilities \cite{sun2020distill,geirhos2018generalisation,qiu2020pre}. Therefore, the human brain, as the only example of the realization of intelligence, has once again attracted the attention of computer scientists, and the development of brain-like intelligence by drawing on the neural and cognitive behavior mechanisms of the human brain has become a hotspot at the forefront of research worldwide.

Simultaneously, cognitive science, including cognitive neuroscience and cognitive psychology, has developed rapidly. Noninvasive and real-time monitoring of language processing in the brain has become possible. Through various sophisticated experimental designs, researchers in the field of cognitive science have made valuable discoveries regarding the neurological basis of language processing \cite{jin2020low,luo2020cortical,sheng2019cortical,lin2018neural,lin2020dissociating}. Traditional cognitive science experiments rely on strict experimental controls, leading to significant deficiencies in terms of ecology and globality. However, in recent years, an increasing number of studies have begun to adopt high ecological paradigms and use advanced data-analysis methods to analyze the information-processing mechanism of the human brain under high ecological validity tasks. In terms of language research, researchers have gradually begun to use computational modeling methods to study the language understanding process of the human brain under experimental stimulation conditions of natural texts \cite{sun2019towards,sun2020neural,shain2020fmri,wang2020fine,toneva2020modeling,deniz2019representation,jain2018incorporating,anderson2017predicting}.

With the rapid accumulation of interdisciplinary research in the fields of cognitive science and computer science in recent years, researchers have begun to summarize the changes that computer science methods, especially deep learning models, can contribute to research in the field of cognitive science \cite{richards2019deep,cichy2019deep,kriegeskorte2015deep,lindsay2021con} and how the conclusions of cognitive science discoveries can help in building artificial intelligence models \cite{lake2017building,poo2018towards,wang2018investigating,cope2018abstract}. The above work explored the possibility of combining the two fields at the macro level but did not discuss how to combine the two fields to carry out work on subdivided issues. To provide a reference for cognitive and computer scientists to conduct interdisciplinary research in the direction of language understanding, this paper summarizes and looks forward to the existing interdisciplinary research work on language understanding.

In summary, in terms of language comprehension, computational models can help cognitive scientists to quantitatively study and model the brain while understanding brain mechanisms can help computer scientists build smarter language and computation models. Therefore, to promote a new round of development in human and machine language understanding research, it is imperative to conduct interdisciplinary research that combines cognitive and computer science. 

The following sections first introduce the definitions, main research issues, research status, and research methods of human language cognition (Section 2) and machine language computation (Section 3) as well as the limitations of existing research. We then compare the main ideas and concepts of language cognition and language computation (Section 4) and analyze the similarities and differences between the two at different levels. Section 5 summarizes the existing work on combining language cognition and computation. On this basis, the limitations of the existing combination methods are analyzed, and feasible future research directions are proposed (Section 6). Finally, we present the main conclusions of this study (Section 7).

\section{Research on language cognition} 
\subsection{Definition of language cognition}
The language cognition mentioned in this article refers to the human brain's understanding of language. Specifically, it refers to the process of extracting abstract symbolic information from auditory, visual, and other sensory information when an individual receives information, such as speech and text. Language cognition is a complex process with varying structures and mechanisms of different levels. Moreover, the brain networks on which they depend are also very complex. For example, in the process of speech comprehension, the auditory system must encode the basic acoustic features of speech and then follow multiple steps, such as vocabulary recognition, syntax construction, and semantic analysis, before finally realizing language comprehension.

\subsection{Main research questions} 
Language is a complex sequence, and the human brain is a complex system. This makes it very challenging to study the language-processing mechanism of the human brain. On one hand, a language contains units of different sizes; which language unit should be used as the starting point? Does the brain use a certain unit as the most important unit in language processing? What about the core processing unit? This is the first research question presented below.

On the other hand, information processing in the brain is very complicated. Different types of information are processed by different brain areas in a certain order. Research on the related brain regions and time courses of language processing is summarized below in the second and third research questions, respectively.

Ultimately, both the observation of the brain area and the processing timing are only phenomenological descriptions of language processing in the brain. What cognitive and computational mechanisms underlie these phenomena? This is the fourth research question introduced below.

\begin{enumerate}
\item Units and dimensions of language cognition

Linguists have defined many language units of different sizes and types such as phonemes, syllables, morphemes, words, phrases, and sentences. A key concern in the study of language cognition is whether these language units are merely concepts proposed by linguists for the convenience of research or truly processing units on which the brain relies for language understanding. In the normal process of language understanding, what type of language unit does the brain analyze? Does the brain construct different neural representations for different types of language information (such as phonetics, grammar, and semantics)? These are the concerns of language cognition research. For example, Liberman et al. \cite{liberman1967perception} believe that the phoneme is the basic unit of speech processing and that phoneme recognition is the function of the brain motor system. However, Greenberg \cite{greenberg1999speaking} and Hickok et al. \cite{hickok2007cortical} hold that the syllable is the more central processing unit and that phoneme processing is possible only for certain tasks. For another example, Townsend et al. \cite{townsend2001sentence} believe that large phrases or even sentences are the basic units of semantic understanding, but many connectionists think that words are the basic processing units of the brain and that phrase and sentence structures have little effect on the brain's language processing \cite{frank2012hierarchical}.

\item Brain networks that localize different types of language information

Language is a function of the brain; but which parts of the brain are crucial for this function? Neuroscience research has found that the brain can be divided structurally and functionally, and the earliest evidence for functional division of the brain comes from studies of aphasia, which will be introduced in the next section. Aphasia research and modern neuroimaging research have found that language is not a single function but includes many functional modules \cite{kay1996psycholinguistic,blank2016syntactic,pylkkanen2020neural}. Therefore, current language cognition research pays more attention to the brain network involved in locating specific functional modules.

\item Time course and control of language information processing

What is the processing order of different modules for language understanding in the brain? For example, will the brain parse the grammatical structure first and then process the meaning \cite{friederici2002towards}? How long does it take to process each step? Can different features in a vocabulary be identified for a long time \cite{laszlo2014never}? Are the steps and sequences of brain processing language automatic and invariable, or do they require the influence and regulation of cognitive functions such as attention and working memory \cite{fodor1983modularity}? These are also the focus of research on language cognition.

\item Neural coding and computational mechanism of language information

Studies of brain regions and time courses have focused on describing language processing phenomenologically. How do these phenomena arise? From a computing point of view, what are the "data structures" for computing in the brain, and what algorithms are used to operate these data structures? Research on processing mechanisms will inevitably involve mathematical models, which also presents difficulty. At present, one type of model seeks to directly explain the neural response of the brain \cite{mitchell2008predicting,huth2016natural,ding2012emergence}, and the other type simulates language behavior, such as by simulating the language-acquisition process \cite{marslen1978sentence}.

\end{enumerate}

\subsection{Development of language cognition} 
Early research focused primarily on aphasia. These studies emerged around the middle of the nineteenth century and mainly analyzed the relationship between brain damage and language behavior in patients. In the past 50 years, the maturity of technologies, including electroencephalogram (EEG), magnetoencephalography (MEG), positron emission computed tomography (PET), and functional magnetic resonance imaging (fMRI), has provided powerful tools for studying the language function of the normal brain. Studies based on behavioral or neuroscience experiments have achieved many results. Combined with the research questions in the previous section, we introduce four studies as examples. 

The first study focused on units of language comprehension. An extreme view is that sentences (or large phrases) form the basis for processing. In this unit, the process of listening or reading a sentence is simply a process of acquiring information, and the information is processed and integrated when it reaches the boundary of the sentence. Another extreme perspective is that language processing is carried out in real time; that is, the brain will process the current information at every moment. Thus, information obtained at all times can be fully processed, and there is no need for centralized processing at some important language boundaries. Evidence for the first view is that language understanding is heavily dependent on context; George Miller found that, in noisy environments, the same word can be better recognized if it is placed in a sentence \cite{miller1951intelligibility}. 

As another example, in the process of listening to an article, if the article is suddenly interrupted and the listener is asked to recall what he heard before, the listener can only accurately recall the vocabulary in the current sentence \cite{jarvella1971syntactic}. The second view is supported by considerable evidence. For example, if you play a voice and ask people to read along, some people can follow at a speed of approximately 300 ms. This means that the voice spoken to the reader is only approximately one word slower than the voice heard. In this case, if you heard the word "tomorrane," but, according to the contextual information, the word should mean "tomorrow," you would have a higher probability of saying "tomorrow" instead of "tomorrane." However, when contextual information is lacking, the reader will not perform this type of correction \cite{marslen1975sentence}. In this case, the reader integrated the above information in real time rather than waiting to process it until the end of the sentence. 

Another study found that, if a person saw four objects on a screen (such as a horse, an apple, a table, and a newspaper), when they heard "this kid is riding," the person's gaze would often fall to the "horse” object. This indicates that people's language processing is predictive and that people instantly generate expectations based on the above. Combining these two perspectives, one can argue that the brain performs both immediate predictive processing and additional integration at sentence or phrase boundaries.

The second study concerned the modules included in language processing and the corresponding regions or networks in the brain. Early research on aphasia found that, when certain brain regions are damaged by trauma or disease, language function is impaired. More importantly, language is not a single function but a complex system of functions in different brain regions. For example, patients with Broca's aphasia cannot produce language but can understand it, and patients with Wernicke's aphasia can speak language but cannot understand it. These two aphasias suggest that language production and comprehension are in separate brain areas and, therefore, can be selectively impaired.

More detailed studies have found that some patients have impaired recognition of nouns but preserved recognition of verbs while others have the opposite, suggesting that verbs and nouns are processed differently in the brain \cite{caramazza1991lexical,silveri1997noun}. Similarly, some patients have an impaired ability to distinguish phonemes (such as being unable to distinguish /ba/ and /da/) but normal auditory word comprehension while others have the opposite, which shows that phoneme discrimination and auditory word recognition involve different brain regions \cite{hickok2007cortical}. All these phenomena indicate that language comprehension involves many modules; thus, damaging specific brain areas affects only part of language function. 

Studies based on fMRI methods, which observe the activation of different brain regions in processing different information or performing different tasks in people with typical brain functioning, have also revealed this functional division. Moreover, they have found that different brain areas are activated by grammatical and semantic processing \cite{skeide2014syntax,friederici2011brain,pylkkanen2019neural}, and different brain regions are activated by different word categories (such as tools versus seeing animals) \cite{binder2009semantic, chen2016representation}. In general, aphasia studies have found that damage to some key brain areas can affect a certain function; however, MRI studies have generally found that this function actually involves a more widespread brain network. For example, traditional aphasia studies generally hold that temporal lobe damage is more likely to cause noun comprehension problems, and frontal lobe damage is more likely to cause verb comprehension problems. However, recent MRI studies have shown that processing verbs and nouns involves very complex brain networks in which internal connection properties are related to the processing of two types of words \cite{yang2017dissociable,binder2016toward}.

The third study examined vocabulary recognition and processing. Numerous studies have shown that word recognition is a parallel process. For example, when hearing the English syllable "/kæp/," it is generally believed that the brain will activate all the words at the beginning of this syllable in parallel, such as cap, captain, and caption. Among these words, those with a higher word frequency or that are more consistent with the context have stronger activation. How do psychologists conclude that many words are activated simultaneously? Classic experiments used the cross-modal priming effect. These experiments found that, after hearing "/kæp/," people recognize visual presentation of words including "captains" and "captions" faster than after hearing other syllables (such as "/da/"). Moreover, after hearing a word, vocabulary related to the semantics of the word is activated. For example, after hearing "captains," people recognize words like "ships" more quickly  \cite{cutler2012native}. 

From the perspective of neuroscience, vocabulary induces an EEG response N400 with a latency of approximately 400 ms, and the amplitude of N400 is closely related to word frequency and the previous context. For example, the amplitude of N400 induced by "ship" depends on the preceding word. If the preceding word is "captain," the amplitude of N400 induced by "ship" is relatively small; if the preceding word is an irrelevant word (such as "apple”), the amplitude of N400 induced by “ship” is relatively large \cite{kutas2000electrophysiology}. 

Vocabulary recognition is generally believed to be a parallel process. After hearing part of the vocabulary information, a large amount of vocabulary is activated; however, as the information increases, the vocabulary that does not match the new information is suppressed until the brain finally determines a possible vocabulary \cite{cutler2012native}. The brain processes possible words in parallel because language is full of ambiguity, and information is presented very quickly. If you wait until all ambiguity is resolved before starting processing, not only may the reaction speed be too slow, but the information that has already exceeded the brain’s working memory capacity may be forgotten. Similar problems are more common in sentence processing, such as the sentence "The horse raced past the fence fell.” Before seeing the last word, we will think that "raced" is the predicate verb of the sentence, but after seeing the last word, we will find that the previous understanding was wrong; "fell" is the predicate verb of the sentence. Sentences that contain ambiguity so that the analysis of sentence structure changes during the comprehension process are called "garden path" sentences. At present, some theories suggest that the human brain will also construct a variety of possible sentence structures at the same time and then continue to screen, but others suggest that the brain will first construct the most likely sentence structure and reanalyze if the structure is found to be wrong.

The fourth study focuses on speech understanding in complex environments. In the 1950s, British scientist Colin Cherry discovered that attention plays a crucial role in speech understanding in complex environments \cite{cherry1953some}. Cherry's and subsequent studies found that, if two different speeches (speeches from different speakers or different spatial orientations) were played simultaneously in the experiment and the listener was asked to focus on one of the speeches, they could understand the speech that they paid attention to very well but could not recall the content of the other speech they did not pay attention to afterwards. Psychologists also quantitatively analyze the impact of various factors on speech recognition, such as measuring the speech recognition rate under different noise intensities, and then draw the psychological curve of the speech recognition rate changing with noise intensity. American scientist George Miller found that speech recognition rate is related not only to noise intensity and listener attention but also to the listener's prior language knowledge \cite{miller1951intelligibility}. In noisy environments, humans can recognize grammatical sentences better than random word strings. 

Cognitive neuroscience experiments have further shown that both attention and prior knowledge can directly regulate the processing of the acoustic features of speech in the auditory cortex. Under this mediation, the neural activity of the auditory cortex mainly encodes the attended speech. These studies demonstrate the important roles of attention and prior knowledge in language comprehension.

The research on the above four aspects shows that predictive processing, language structure processing, parallel processing, attention, and prior knowledge are all important characteristics of human language cognition.

\subsection{Research methods} 
Studies in cognitive and life sciences can be divided into hypothesis- and data-driven research. Accordingly, linguistic studies can also be roughly divided into these two types of research.

\begin{enumerate}
\item Hypothesis-driven research

Most language cognition research is hypothesis-driven; that is, researchers will clarify the hypothesis to be verified by the experiment (generally referred to as H1) and its opposite hypothesis (H0) before the experiment. They will also clarify how the experimental results are consistent with the expectations of H1 or H0 unanimously. If the final experimental result is consistent with the expectation of H1 and inconsistent with the expectation of H0, then H0 is falsified (that is, the findings support H1). In contrast, if the final experimental result is consistent with the expectation of H0 and inconsistent with the expectation of H1, then H1 is proven false. 

For example, suppose researchers wanted to test the hypothesis that the auditory cortex encodes not only the acoustic features of speech but also phonemic information. Based on this hypothesis, the researchers designed experiments to distinguish acoustic features from phonemic features and then analyzed whether the latter affected the responses of the auditory cortex. Two specific examples below illustrate hypothesis-driven research.

As mentioned above, the unit of brain processing language is a controversial issue. The hypothesis proposed by one study is (H1) that the brain can encode multiple levels of language units in parallel and that the neural activity encoding a language unit should be synchronized in time with the unit; that is, when a language unit appears, the neural activity also occurs, and when a language unit ends, so does the corresponding neural activity. The counter hypothesis (H0) is that the brain processes only according to a single level (such as words) or that different levels of language units do not show neural responses synchronized with language units. If this assumption holds, then the update rates of neural activity encoding language units of different sizes, such as syllables, words, phrases, and sentences, will be different. For instance, if there are four syllables in speech per second, the response to the syllables will also change four times per second. If every two syllables are combined into a phrase and every two phrases form a sentence, then the neural response of words and phrases should be updated two times per second, and the neural response of sentences should be updated one time per second. Therefore, the researchers designed the experiment according to the above ideas and found that the MEG/EEG response of the human brain in the process of listening to speech does contain 4 Hz, 2 Hz, and 1 Hz components, corresponding to the neural responses of hypothetical syllables, phrases, and sentences \cite{ding2016cortical}. This experimental result supports H1.

Another example is the study of word meanings. The semantics of a word can be learned in two ways. One way is to directly establish the relationship between the word and the objective entity it refers to, such as seeing the fruit "guava" and being told it is a "guava." This learning method directly establishes the relationship between the language symbol "guava" and the sensory characteristics (visual, taste, etc.) of the object it refers to. Another way is to describe the meaning of the acquired vocabulary through words, such as reading in the dictionary, "Guava is a plant of the myrtle family, and the fruit is edible." The hypothesis (H1) here is that the semantics acquired through the above two methods are encoded in different regions of the brain. The opposite hypothesis (H0) is that the representation of a word in the brain is independent of the acquisition pathway. To distinguish between these two modes of acquisition, the study compared the processing of colors by sighted people and congenitally blind people; both groups can acquire color concepts through language, but only sighted people can directly establish color words and visual correspondence between color information. The study found that some brain regions encode color in the same way in both groups, but others encode color in a way that is only present in sighted people. This result also supports H1 hypothesis \cite{wang2020two}.

\item Data-driven research

Hypothesis-driven research is often highly targeted research--that is, experiments specifically designed to test a particular hypothesis. The opposite of hypothesis-driven research is data-driven research. Data-driven research is exploratory; it does not put forward a hypothesis first but explores possible results by collecting experimental data. The purpose of hypothesis-driven research experiments is clear, so it is easier to obtain stable results. For example, to verify the hypothesis that neural activity and hierarchical language structure are synchronized in the first experiment above, a constant rate was used to play speech to simplify data analysis (the researchers only needed to analyze the response of a specific frequency in the frequency spectrum). Moreover, to distinguish the two methods of acquiring word meaning in the second experiment, two groups of people were selected for comparison. 

However, it is often difficult to strictly distinguish between hypothesis- and data-driven research. Without exploratory research, it is difficult to formulate a hypothesis; without a hypothesis, it is difficult to determine which aspect of the data to analyze. For example, the above two hypothesis-driven studies also contain data-driven components. The first study did not determine which MEG/EEG channels can obtain responses, and the second did not assume in advance the brain region in which the experimental phenomenon would be found.

\end{enumerate}

\subsection{Limitations of existing research} 
Language cognition research initially revealed some patterns of human language understanding, but much more is needed to truly analyze the mechanism of language understanding in the human brain. Currently, the main problem is that, in theory, the existing research focuses on the qualitative explanation of local problems and relies on small samples and strict experimental control, which leads to a lack of ecological and global research conclusions. The overview is as follows.

\begin{enumerate}
    \item Lack of discussion on the quantitative mechanism
    
    Most cognitive science research is described at the phenomenon level, and even the discussion of the mechanism is often qualitative and subjective. For example, as mentioned before, when the brain processes a word, it produces an EEG response of N400. Many of studies have investigated this response, clarifying how the previous contextual information of various properties affects the magnitude of the N400. However, these studies only show that the previous contextual information can affect the N400 response and do not answer what computational mechanism this effect reflects. Cognitive science literature discusses multiple mechanisms for the generation of N400. One hypothesis is that N400 represents the current word that can be predicted by the brain; that is, words that can be predicted produce smaller N400. Another hypothesis suggests that N400 represents the ease of integration of a current word with previous context; that is, N400 is smaller if it is easy to integrate. However, these hypotheses are qualitative language descriptions, and it is not clear how the brain's prediction and integration reflect the computational mechanism.
    
    \item Targeting specific linguistic phenomena
    
    Cognitive science experiments often use strictly controlled experimental designs to study specific, even very detailed language phenomena. Due to the strict control of experimental variables, the corpus in the experiment tends to be consistent, so the experimental conclusions are likely to be applicable only to the highly consistent corpus involved in the experiment. Poeppel and Embick \cite{poeppel2017defining} identified a mismatch of research scales between linguistics and neuroscience research. Linguistics is often concerned with very fine-grained issues (such as the usage of a word and how the syntactic structure of a sentence should be divided) while neuroscience is concerned with relatively macro issues (such as which part of the brain processes grammar). However, even neurolinguistics studies often only use a relatively consistent and typical corpus for research, so the universality of the conclusions is not strong.
    
    \item Research conclusions are difficult to integrate
    
    Closely related to the previous study, tightly controlled experiments lead to fragmentation of research with one study only concerned with one particular linguistic phenomenon. If each study focuses on one linguistic phenomenon and language contains a limited number of linguistic phenomena, then an overall conclusion can be drawn by integrating different local studies. However, because the language is too complex to be uniformly divided into several basic phenomena and the experimental methods are too diverse, it is very difficult to integrate various research conclusions. For instance, cognitive experimental studies have found that different types of language materials can activate different brain regions and induce different types of EEG responses. However, if these studies are combined, can they tell us how the brain understands even a simple sentence step-by-step? In fact, they cannot.
    
\end{enumerate}

\section{Research on language computation} 
\subsection{Definition of language computation}
The language computation mentioned in this article refers to the process of machine understanding of language. Taking Chinese as an example, the process of language computation includes the recognition and representation of characters, the structure and semantic analysis of texts (including words, phrases, sentences, and discourses), and the analysis of the association between text symbols and the external world and finally achieves the goal of enabling machines to understand language. The language computation we refer to below can be compared with the basic research problems in natural language processing (also called natural language analysis), such as lexical, syntactic, semantic and discourse analysis, knowledge representation, and computing, and does not involve application technology research. 

\subsection{Main research questions}
To make a machine understand natural language, we must first encode the information in a language into a form that can be processed by the computer, which is called the text representation task. To further analyze the information in a text, it is necessary to analyze its structure and semantic information--that is, to perform structural analysis and semantic analysis tasks. Thus far, the machine has known the relationship between different language symbols. It is necessary to further associate language symbols with the external world and knowledge to understand natural languages like humans. The main research questions are as follows:

\begin{enumerate}
    \item Text representation method
    
    Language is composed of small elements hierarchically and recursively, which in turn form words, phrases, sentences, and discourses. In language communication, word is the most basic semantic unit, and the combination of words needs to be based on specific rules. These limited rules can combine different concepts to construct endless text units. How to  represent lexical semantics, such as by using symbols \cite{chomsky2014aspects}, functions \cite{vilnis2014word}, vectors \cite{mikolov2013efficient}, and tensors \cite{smolensky1990tensor}? How to build efficient lexical semantics learning methods\cite{palatucci2009zero,wang2018associative,peters2019knowledge}? How to combine lexical meaning to form the meaning of larger-grained text units \cite{mitchell2010composition,ling2015finding,wang2017comparison,wang2018empirical}? These are key issues in linguistic computing.
    
    \item Structural Analysis Methods
    
    Structural analysis is generally divided into syntactic structure analysis and discourse structure analysis tasks. Among them, syntactic structure analysis studies the combination and dependence relationship between words in a sentence, and discourse structure analysis studies the combination and dependence relationship between sentences in a paragraph or discourse. These two types of analysis can resolve the ambiguity of the structure in the input text, analyze the internal structure of the input text, and provide structural information for the semantic analysis of the text, which is considered to be an important part of language understanding \cite{shen2018straight,li2018seq2seq,liu2018discourse}. The main research issues in this direction include how to design or select formal rules for grammars and how to design automatic analysis algorithms. Representative of this kind of work are the rule-based phrase structure analysis method proposed by Klein et al. \cite{klein2003accurate} and the dependency structure analysis method based on neural networks proposed by Chen et al. \cite{chen2014fast}.
    
    \item Semantic analysis method
    
    For different language units, the task of semantic analysis is different. For word, semantic analysis focuses on how to disambiguate the meaning of words and how to identify the semantic relationship between words (including antonyms, synonyms, part-whole and event relations, etc.); for sentences, semantic analysis includes semantic role labeling, semantic parsing, calculation of semantic similarity between texts, and identification of implication relations; for discourses, semantic analysis includes how to resolve references and identify inter sentence relations in texts. The identification and calculation of the above-mentioned semantic information and semantic relationship is the basis for understanding the meaning of a text. It is also a difficult problem in language computation to build an efficient semantic analysis method \cite{navigli2009word,kamath2018survey}.
    
    \item Knowledge representation and symbol association method
    
    Knowledge in this paper refers to world knowledge, historical knowledge, commonsense knowledge, and professional knowledge of various disciplines. Knowledge representation is a description of knowledge. The current model represents knowledge in the form of symbols \cite{miller1995wordnet} or distributed vectors \cite{socher2013reasoning} and realizes the association between language symbols and knowledge through retrieval or mapping to a unified representation space. Among them, how to design the encoding form of knowledge and automatically learn the representation of knowledge is the key to this type of research \cite{wang2017knowledge,guo2018knowledge}. In addition, no matter it is a human or a machine, to understand the meaning encoded in language symbols, it is necessary to associate it with world knowledge. Otherwise, as the "Chinese room" described, the people in the room do not know Chinese and cannot truly understand the received Chinese information, but he can make Chinese native speakers think that he can speak Chinese fluently, creating an intelligent impression \cite{harnad2005searle}. Therefore, how to associate language symbols and knowledge is also the core issue of language computation research.
\end{enumerate}

\subsection{Developments of language cognition}

Language is a serialized and structured symbolic expression. How to represent the meaning of text and automatically analyze its semantics and structure is a crucial step in research on language computation. Moreover, this has always been a major challenge in machine language understanding. Almost all natural language processing tasks, such as machine translation, question answering, and dialogue systems, rely on semantic representation and computation of input language sequences.

Amidst the decades of the development of natural language processing, text-representation methods have undergone a systematic transformation from discrete symbol representation to continuous vector representation. With discrete symbol representation, words are regarded as discrete symbols, and each word can be expressed as a one-hot vector whose dimensions are equal to the size of the vocabulary, where one dimension is 1 and the other dimensions are 0. In this representation system, sentences and discourses are usually represented by a bag-of-words model. 

In 1954, Harris proposed the concept of a bag of words in the article "Distributional Structure." In the following decades, the bag of words model has been the mainstream model of text representation \cite{harris1954distributional}. This text representation method, based on discrete symbols, can only use string matching to extract features and calculate the similarity between language units, which easily leads to data sparsity problems and cannot capture the semantic similarity between words. 

On the other hand, distributed continuous vector representation is convenient for semantic calculation and measurement and can theoretically solve the problem of semantic gaps between words, sentences, and discourses. Harris and Firth proposed and clarified the distributed hypothesis of words in 1954 and 1957, respectively, in which the semantics of a word are determined by its context; that is, words with similar contexts have similar semantics \cite{harris1954distributional,firth1957synopsis}. Matrix decomposition and neural networks are the two main models for learning the distributed vector representations of words. Among these, neural networks have been the mainstream model for learning distributed vector representations in recent years. 

In 2003, Yoshua Bengio et al. \cite{bengio2000neural} proposed a neural network language model that uses a low-dimensional continuous real number vector to represent each word and learns an n-gram grammar model based on this, marking the beginning of distributed text representation. Tomas Mikolov et al. \cite{mikolov2013efficient} proposed the Word2Vec method, including two models of CBOW and Skip-gram in 2013, which greatly simplifies the distributed vector learning method of words so that it can make full use of massive unlabeled text data to learn words efficiently. 

In 2017, the transformer model proposed by Google \cite{vaswani2017attention} combined the semantics of vocabulary more efficiently through pairwise calculations between words to obtain a semantic representation of text. Since then, largescale pre-training models based on transformer architectures, such as BERT \cite{kenton2019bert}, TransformerXL \cite{dai2019transformer}, GPT3 \cite{brown2020language}, PaLM \cite{chowdhery2022palm}, and ChatGPT \footnote{\url{https://openai.com/blog/chatgpt/}}, have been developed for various language processing applications, further establishing the dominance of distributed vector representation.

The distributed text representation model greatly facilitates the representation and calculation of natural language, thus becoming the cornerstone of deep learning applied to natural language processing tasks and further promoting the breakthrough development of applications such as text understanding and machine translation. However, existing methods lack the modeling of fine-grained text semantics and structural information and cannot effectively deal with linguistic phenomena such as lexical ambiguity, antonyms, extended meanings, and structural ambiguities of sentences and texts \cite{shwartz2019still,loureiro2021analysis,faruqui2014improving}. On the other hand, semantic, syntactic, and discourse analysis methods study how to represent the meaning, combination, and dependency of language units and provide structural information for text representation, which may be helpful for existing text representation models.

Semantic parsing is the most representative task in semantic analysis, which studies how to parse natural language into a complete semantic representation (such as logical expressions) that can be recognized or calculated by machines. Specifically, it includes how to obtain the semantics of words in a sentence, the semantic relationship between words, and how to combine word representations into sentence semantic representations \cite{berant2013semantic}. Similar to most natural language processing tasks, semantic analysis has undergone a shift from rule-based, statistical, and neural-network approaches. Although the performance of the neural-network-based semantic analysis model has made great progress, this method still strongly relies on a large amount of manually labeled data and cannot truly realize semantic understanding by learning statistical laws for prediction. 

In addition to the abovementioned semantic analysis, understanding the meaning of a text is inseparable from the analysis of its structure. At the sentence level, structural analysis refers to the analysis of the input word sequence in the syntactic structure of a grammatical sentence, which is mainly divided into phrase and dependency structure analysis. Phrase structure analysis analyzes sentences in a hierarchical phrase structure tree based on phrase structure grammar, which is part of the theory of language transformation and generation created by Chomsky \cite{chomsky2009syntactic} in 1957. Dependency structure grammar is mainly used to describe the semantic dependencies between words and was first proposed by Tesniere in 1959 \cite{tesniere1959elements}. 

At the discourse level, structural analysis studies the combination and dependency relationship between sentences in paragraphs or discourses with the goal of analyzing discourse structure as a whole to understand discourse meaning. Computational linguists generally believe that syntax and discourse analyses are necessary to realize language understanding. Only by analyzing the structure of the text correctly can we understand its meaning. However, the results of various language processing tasks in recent years have shown that explicitly modeling text structures does not significantly improve model performance \cite{bowman2015tree}. This leads researchers to rethink the role of text structure analysis, whether existing language models can implicitly learn certain syntactic rules, and whether artificially defined syntactic structures are necessary for machine-language understanding.

In addition to expressing and analyzing shallow meaning units such as words, phrases, and sentences, language understanding also requires higher-level expressive abilities, such as the connection with "knowledge," the ability to perceive context, and reasoning. Traditional knowledge representation methods use a computational framework based on logical symbols, such as a logical reasoning language based on first-order predicate knowledge representation represented by Prolog and a probabilistic graphical model represented by a Markov logic network. These methods are good at precise reasoning and computation but lack generalization and approximate semantic computation capabilities under incomplete knowledge. In recent years, with the development of deep learning, knowledge representation has gradually adopted neural-network-based methods to learn the representation of entities and relations. This method based on representation learning has better generalization performance and is more conducive to the widespread uncertainty calculation problems in information processing, but it cannot perform precise semantic calculations and reasoning \cite{ji2021survey}. Therefore, the combination of precise semantic computing based on symbolic logic and approximate semantic computing based on representation learning is the future research trend of knowledge representation methods.

\subsection{Research methods} 
Consistent with most research directions in the field of computer science, natural language methods are mainly divided into two camps: rationalist and empiricist (or a data-driven approach). The rule-based approach is representative of the rationalist approach, which holds that a large part of people's language knowledge is innate and determined by genetics. Data-driven approaches include statistical machine learning methods and neural network methods (or deep learning methods). This method assumes that the complex language structure of humans can be learned through acquired training. 

\begin{enumerate}
    \item Rule-based approach
    
    Since natural language is essentially a symbolic system produced by human society due to the need for communication, its rules and reasoning features are distinctive, so the early research on natural language processing first adopted the rule method. This method obtains the understanding of people's language ability through the study of some representative sentences or language phenomena and summarizes the laws of language use to analyze and infer the expected results of the test samples. The rationale for rule-based approaches is presented here using the application of context-free grammars to the problem of syntactic parsing as an example. For example, we have the following context-free grammar:
    
        $G = (V_n, V_t, S, P)$ \\
        $V_n = {S, NP, VP, N, V, Det}$ \\
        $V_t =$ {one, kid, chases, a, cat} \\
        $S = S$ \\
        $P:${ $S \rightarrow NP \: VP$, $NP \rightarrow Det \: N$, $VP \rightarrow V \: NP$,  $Det \rightarrow$ one, $ Det \rightarrow$ a, $N \rightarrow$ kid, $N \rightarrow$ cat, $V \rightarrow$ chases}

    Based on the above rules, the syntax tree can be constructed by the top-down or bottom-up analysis method. Here, we use the top-down analysis method as an example for illustration. First, start from the initial symbol $S$, search from top to bottom, select the applicable rule in the grammar $P$ to replace the search target, match the word in the sentence with the right part of the grammar rule, and erase the word if the match is successful. Then, the search continues on the remaining part of the input sentence until the end of the sentence. Specifically, in the above example, the search steps are: 

    \begin{enumerate}
        \item Select rule $S \rightarrow NP \: VP$, no matching word, left sentence "one kid chases a cat".
        \item Select rule $NP \rightarrow Det \: N$, no matching word, left sentence "one kid chases a cat".
        \item Select rule $Det \rightarrow$ one, matching word "one", left sentence "kid chases a cat".
        \item Select rule $N \rightarrow$ kid, matching word "kid", left sentence "chases a cat".
        \item Select rule $VP \rightarrow V \: NP$, no matching word, left sentence "chases a cat".
        \item Select rule $V \rightarrow$ chases, matching word "chases", left sentence "a cat".
        \item Select rule $NP \rightarrow Det \: N$, no matching word "a cat".
        \item Select rule $Det \rightarrow$ a, matching word "a", left sentence "cat".
        \item Select rule $N \rightarrow$ cat, matching word "cat", no left sentence.
    \end{enumerate}

    According to such a search process, we can express the syntax tree of the sentence "a child chasing a cat", as shown in Figure \ref{fig2}.

    \begin{figure}[htbp] \centering
    \includegraphics[width=0.3\textwidth]{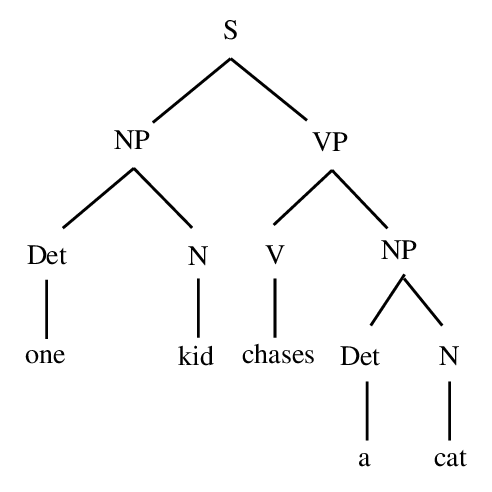}\\
    \caption{Example of a syntax tree}\label{fig2}
    \end{figure}

    \item Data-driven approach
    
    Human language is not a formal language for rules and patterns often exist implicitly in the language, making the rules difficult. In addition, the complexity of natural language makes it difficult for rules to cover all linguistic phenomena without conflicts. The data-driven method saves the burden of manually compiling rules and automatically generates features or evaluates the weight of features in model generation, which has good robustness. Generally, data-driven methods include statistical methods and neural network methods, and the following two methods are introduced.
    
    Statistical-based language computation methods use large-scale language data and often need artificial help (labeling data and screening features, etc.). They use statistical methods to discover the law of language use and its probability which are then be used to calculate the possible outputs of testing data. Here, we take the language model as an example to introduce the basic principles of statistical methods in which the representative one is the n-gram model.
    
    The goal of the language model is to calculate the probability of a string appearing as a sentence, that is, to calculate the product of the probabilities of each word in the sentence $s$ in different historical situations: $P (s) = p(w_1) \times p(w_2| w_1) \times p(w_3|w_1 w_2) ... p(w_l|w_1 ... w_{l-1})$. Among them, the probability of generating the $i$-th word $w_i$ is determined by the generated $i-1$ word $w_1 w_2 ... w_{i-1}$. When the above method is simplified and only the previous $n-1$ words are used to predict the next vocabulary, it is an n-gram model. For example, the bigram model calculates the probability $p$(a child chasing a cat) = $p$(one$|$⟨bos⟩)$\times p$(child$|$one)$\times p$(chasing$|$kid)$\times p$(one$|$ (cat$|$one). 
    Among them, 〈bos〉is the start character. Generally, maximum likelihood estimation is adopted to calculate the conditional probability of $P(w_i|w_{i-1}) = \dfrac{c(w_i|w_{i-1})}{\sum_{w_i} c(w_i, w_{i-1})}$.
    
    Methods based on neural networks mainly study how to design task-related neural network structures and optimize neural network parameters. Here, we take the text representation problem as an example and use the recurrent neural network model as an example to introduce the basic principles of neural network-based methods. A recurrent neural network is a kind of neural network that is good at processing sequence information and long-distance dependencies. For example, the corpus has the following content: "A child chases a cat ...". To represent the above text, the sentences in the text need to be sequentially input into the cyclic neural network. As shown in Figure \ref{fig3}, the cyclic neural network model processes each vocabulary in the sentence in turn, encodes the corresponding vocabulary wt into a real-valued vector $x_t$ at time $t$, and combines it with the hidden layer vector $h_{t-1}$ generated at the previous time to obtain the hidden layer vector at this moment:
    $h_t:= \sigma (W_xh_{xt} + W_hh h_{t-1} + b)$,

    \begin{figure}[htbp] \centering
    \includegraphics[width=0.6\textwidth]{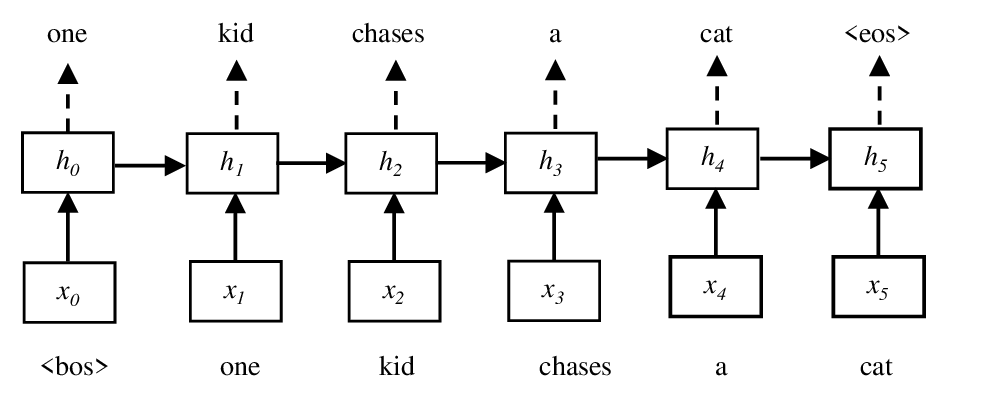}\\
    \caption{Example of a recurrent neural network}\label{fig3}
    \end{figure}
    
    Among them, the symbol ":=" means "defined as"; $\sigma(z) = 1/(1 + exp(-z))$; $W_{xh}$, $W_{hh}$ and $b$ are the model parameters. The objective function of the model is to maximize the probability of predicting the following vocabulary above. The commonly used objective function is to minimize the cross-entropy loss of the model: Loss$=- \sum w_t log y_t$. $w_t$ is the real following vocabulary, and $y_t$ is the following vocabulary predicted by the hidden layer vector $h_t: y_t = Softmax(V_{hy} h_t + c)$, where $V_{hy}$ and $c$ are model parameters.
    
    The above objective function is used to learn the parameters of the model on large-scale unlabeled texts so that the model can predict the following words through the above words. In this process, the language model obtains the representation vector of the vocabulary. At the same time, the representation vector of the sentence can be the hidden layer vector obtained at the last moment of the model or the maximum or average pooling result of all hidden layer vectors.

\end{enumerate}

\subsection{Limitations of existing research}
Existing language-computation models are far from being able to understand language similar to humans. The main problem currently faced is that the model structure lacks a theoretical basis and parameter training relies on large-scale computing resources, which can be summarized as the following four points:

\begin{enumerate}
    \item Single textual representations
    
    Existing text representation models do not distinguish between different types of information (such as vocabulary of different parts of speech, text units of different granularities) and uniformly encode them into dense vectors of the same dimension. This encoding method is very efficient when constructing a neural network model, but it ignores the size of different types of texts. Therefore, to encode all information, it is necessary to uniformly use the encoding method with the largest amount of information for all types of texts. This method adopts larger numbers of parameters instead of designing model structures and contains a large number of redundant parameters.
    
    \item Lack of interpretability
    
    Although the deep learning method has greatly improved the performance of numerous tasks in natural language processing, the meaning of the vector dimension cannot be explained. Therefore, it is impossible to analyze what operation each unit in the network performs on the language input. Moreover, it is impossible to give the reason the model obtained the wrong result. This makes the results unreliable and hinders the design of the model structure and further improvement of performance. 

In contrast, if the model is interpretable and can give the reason for reaching a certain conclusion like a human, then the model structure can be improved in a targeted manner, thereby improving the model performance. For example, when judging whether two sentences express the same meaning, the model can give a yes or no conclusion and, at the same time, give which text fragments in the two sentences significantly affect this conclusion. These can be used as the basis for judging the rationality of the model.
    
    \item Lack of ability to learn independently
    
    Existing methods construct training datasets for each task, adopt the "training-test" development method, and cannot directly use new training samples to correct the trained model. Additionally, the model trained in a certain task is difficult to apply to other middle tasks. Different language tasks require the model to be capable of language representation and understanding, so new tasks can use the already-trained model to learn new task-specific information on this basis. Analogous to the self-evolving learning ability of humans--that is, learning from simple tasks and continuously learning new tasks and correcting existing knowledge on this basis--an intelligent system should also be capable of continuous learning to achieve the self-evolution of the model.
    
    \item Rely on largescale single modality training data
    
    Existing language-computation models rely heavily on the quality and scale of training samples, making it hard to handle those words, language structure, and language expressions that have not appeared in the training corpus. In contrast, when humans learn language concepts, they often obtain information from multiple modal samples and can learn a new word or language expression with only a few samples. For example, when learning the concept of "giraffe," we might understand from a text that "it is the tallest mammal in the world, and its distinctive features are long neck" and so on. When viewing a picture of an animal with a "long neck, long legs, body" and "a spotted pattern," we can quickly recognize that it is a "giraffe" even if we have never seen one before. This shows that multiple-modal data complement and verify each other. Thus, it is imperative to develop small-sample learning algorithms by comprehensively utilizing multiple-modal information.
    
\end{enumerate}

\section{Comparison between language cognition and language computation}

The above chapters demonstrate that the processes of human and machine language understanding defined in the fields of language cognition and language computation are similar--that is, starting from the encoding of basic language units, then combining them into larger-grained text units, and finally connecting to the outside world. However, the experimental methods and focus of the research questions in the two fields are quite different. 

Language cognition research primarily adopts the following experimental method: "1) proposing hypothesis; 2) designing experiment; 3) hypothesis verification." Language computation research primarily adopts the method of "1) collecting data; 2) building model; 3) model performance verification." In terms of research issues, language cognition research tends to reveal the reasons behind problems whereas language computation research focuses on methods to solve these problems. Table \ref{table1} summarizes and contrasts the main ideas and concepts of the two fields. Language cognition studies the basic units and dimensions of language understanding, its spatial representation and time course in the brain, and the regulation of world knowledge and cognitive functions on language understanding, emphasizing "what" and "why" questions. Language computation studies how to allow computers to efficiently segment and represent vocabulary, combine vocabulary representations into sentences and discourse representations, and analyze the functions and relationships of each element in a text. The main concern is the question of "how to do it."

\begin{table}[h!]
  \begin{center}
    \caption{Main concepts in language cognition and language computation areas}
    \label{table1}
    \footnotesize
    \tabcolsep 10pt 
    
    \begin{tabular}{l|p{5cm}p{5cm}} 

       & \textbf{Language cognition} &  \textbf{Language computation} \\ \hline
      
      Language processing unit & Language is organized according to different levels of structure (such as morphemes, words, phrases, sentences, discourses, etc.), with different levels of information (phonetic, grammatical, semantic, etc.) & Word is generally used as the basic unit, and they are integrated through language models such as neural networks \\ \hline

      Word representation & The neurons or neural networks related to the word are activated. It is generally believed that the neural representations of multiple words can be activated at the same time & Encoding words into a form that can be processed by computers (such as symbols, vectors,  matrices, graphs, etc.) \\ \hline

      Word form analysis & Morphology information encoding & Remove the affix to get the root (plays $\rightarrow$ play) or transform the complex form of the word into the most basic form (are $\rightarrow$ be) \\ \hline
    
      Information integration & The process of assigning syntactic structure and semantics to the corresponding input words  & Analyze the semantic and syntactic information and their relationships, then combine word representations to form the representation of sentences and discourses through a composition function \\ \hline
    
      Multimodal information & World knowledge (common sense) and language understanding scenarios can affect language processing, which may involve the interaction of brain language networks with sensory and motor networks & Associate language symbols with other modality information \\ \hline
    
      Task effect & Language comprehension is regulated by cognitive functions --- if cognitive functions such as attention and working memory are adjusted through experimental tasks, cognitive neural processing of language will be affected & The parameters of the computational language model are determined by the objective function --- once the objective function of the task changes, the parameters of the computational language model will change accordingly \\ \hline
      
    \end{tabular}
  \end{center}
\end{table}

These differences carry substantial challenges and opportunities to the combination of language cognition and computation. One challenge is that studies on linguistic cognition and computation are in separate fields. Research on language cognition has mainly focused on the cognitive laws behind human language and its neural basis. Owing to the limitations of data collection and analysis methods, such research still focuses on qualitative analysis and exploration of macroscopic laws. It lacks comprehensive and quantitative laws and models that can be applied in practice. 

An example is the representation of lexical semantics. Although many semantic dimensions have been discovered in the field of language cognition, their general neural connections have been revealed, and some theoretical models have been initially established. However, owing to technical limitations, it remains impossible to accurately record the neural activations of a single word. Therefore, it is difficult to reveal the underlying neural representation rules and calculation methods of semantic information directly. 

Similarly, computational linguists have focused on the effectiveness of practical applications when choosing research questions. They often ignore the most essential laws of language; they only consider the performance in downstream tasks when constructing a language-computation model and ignore interpretability and human-like features. Therefore, as a "black box" that lacks human characteristics, the language-computation model is difficult to use in the modeling of human brain language understanding.

To illustrate these challenges further, let us consider the word representation problem as an example. Regarding the problem of lexical (concept) representation in the human brain, Wang et al. \cite{wang2018organizational} examined the representation of abstract words. By constructing two different types of lexical representation (statistical co-occurrence- and semantic feature-based), they tested two classical cognitive theories of abstract concept representations in the brain. These two theories highlight that abstract word representations are expressed in linguistic symbols through contextual associations or semantic features. The experimental results show that both types of lexical representation have significant effects on the brain. The difference is that corpus-based abstract lexical representations are associated with brain areas responsible for advanced language processing whereas semantic feature-based abstract lexical representations present distributed representational features that are associated with multiple brain regions. Therefore, the researchers concluded that the representation of abstract words in the brain is divided into two modes and is responsible for different brain networks. 

This study revealed that the semantic dimension of abstract vocabulary encoded by the brain may include both linguistic and semantic features. This is very important for further analysis of the brain's understanding of language. However, this analysis is too macroscopic to directly guide the construction of computational models. Computer scientists want to understand the specific semantic dimensions and forms of encoding that the brain needs to encode lexical concepts. For example, which neurons in the brain encode the concept of "knowledge"? What information do these neurons represent? What is the relationship between these neurons? What are the rules for their connection? The answers to these questions can directly inspire the development of new lexical representation model architectures. 

In contrast, when the language-computation model encodes the meaning of words, it represents the words as dimension-agnostic real-valued vectors and encodes the information of the words through the relationship between the vectors. The purpose is to allow computers to process language symbols efficiently to complete various language tasks. Although this uninterpretable representation method is crucial for computational models to complete language tasks, it does not reveal the laws of language itself or directly shed light on how the brain encodes lexical concepts. What cognitive scientists want to understand is which features can be used to construct semantic vectors to explain human behavioral data? What information is encoded in the computational model, and what kind of operation is performed to lead to its excellent performance in downstream tasks? Can the calculation process explain the human language-processing mechanism? Which calculation module in the calculation model is essential for language modeling, and is there a corresponding processing module in the human brain?

In summary, the fields of language cognition and language computation differ in their research contents and ways of thinking. However, we believe that such differences can bring new insights into both fields. For example, in the process of language comprehension, the human brain not only combines words into sentences and discourse from the bottom up but also uses cognitive functions such as attention and working memory to regulate the language process from top to bottom. In contrast, language computational modeling is a static process unaffected by the external environment and computes a fixed encoding result for a piece of text. However, if we can learn from the dynamic encoding mechanism of the human brain and integrate modules such as human brain memory and attention to construct a new language-computation model, the model may acquire more general knowledge and be easier to transfer to other tasks. 

As another example, experiments have shown that, in the process of machine language understanding, addition is a very effective combination when integrating vocabulary representations into phrase representations. This automatic learning of language combination rules from data may provide new ideas for the study of the word combination process in the human brain using the underlying calculation method. Additional ideas on combining the two are introduced in Sections 5 and 6.

\section{Convergence of language cognition and language computation} 
Recent years have seen increasing attention to cross-disciplinary research in the fields of cognitive and computer science. The following section introduces related work in the fields of language cognition and computing that inspire and merge with each other.

\subsection{Language cognition experiments using language computation methods}
In recent years, an increasing number of researchers have begun using language computation methods to study the process of understanding human language. This method shows great potential for studying brain representations at the single-word level. Furthermore, it can be used to analyze both traditional experimental data and natural language-processing data. Specifically, this type of method collects neural activity data of words, sentences, or chapters; uses language-computation models to encode experimental stimuli; and uses the encoded stimuli to study the problem of brain language understanding. Such methods typically work as follows.

Mitchell et al. \cite{mitchell2008predicting} published an article in "Science" in 2008 regarding the issue of how the brain represents conceptual semantics. They found that fMRI data of people reading nouns can be modeled using the statistical laws of certain action words. Specifically, they collected fMRI data when research participants read 60 noun stimuli (pictures + lexical texts) and calculated the representation vectors of these 60 nouns by using their co-occurrence with 25 sensory-motor-related representative verbs (e.g., "see," "listen,” "speak," "eat”). These representation vectors were then trained to predict fMRI data using a leave-two-out cross-validation method; each cross-validation predicted fMRI data for two test words and compared them with the real fMRI data as test accuracy. The regression model had a significantly higher classification accuracy than the random value for brain activation patterns evoked by nouns. This suggests that there is a predictable relationship between word representations and fMRI data. Moreover, it provides evidence that the brain represents the semantics of nouns that are significantly dependent on sensorimotor properties. This work has changed the previous experimental paradigm of examining concept representation only through the comparison between semantic categories, opened a new data-driven method for studying a single lexical concept, and inspired many subsequent studies.

Another representative work is an article published in "Nature" by Huth et al. \cite{huth2016natural} in 2016. They used language-computation models to comprehensively study how different semantic information is encoded in the brain. Specifically, they collected fMRI data when the participants listened to more than 2 h of narrative stories (including a total of 10,470 different words) and selected 985 basic words describing different topics in the corpus as different semantic attributes. Then, they constructed a 985-dimensional vector for each word in the stimuli by counting their co-occurrence with the 985 basic words in a large-scale corpus. 

Next, they trained a ridge regression model such that 985-dimensional word vectors predicted each voxel in the fMRI data. Among them, the parameter matrix with the size of "985 × number of voxels" obtained in the model was the brain representation of 985 semantic attributes. The results showed that the brain semantic representations were very similar across participants, and different semantic features were encoded in specific brain regions. In contrast to the study by Mitchell et al. on the representation of 25 sensorimotor attributes in the brain, Huth et al. comprehensively studied the representation mode of 985 semantic features in different voxels of the whole brain for the first time. This once again proved the usefulness of the language-computation model in studying the brain's language understanding.

Regarding the brain's computational mechanism for language processing, Brennan et al. \cite{brennan2016abstract} studied whether the brain uses a linear or hierarchical structure to process sentences. They first collected fMRI data while participants listened to the first chapter of "Alice in Wonderland." They then used a series of grammatical models (including linear and hierarchical grammar models) to calculate the syntactic complexity of each word in the stimulus. Finally, regression analysis was performed between the syntax complexity index and the fMRI data. 

Syntactic complexity can be calculated from the probability of words appearing in a certain grammatical structure. The hypothesis is that, if people use hierarchical structures to comprehend stories, then the complexity indicators calculated using hierarchical grammar should have stronger correlations with the fMRI data compared to the linear grammar model. The results show that the linear effect is widely distributed in the language network of the brain while only a specific area of the left temporal lobe of the brain is responsible for processing hierarchical structure information. This suggests that the temporal lobe processes information in a hierarchical structure when comprehending languages.

Research on language cognition also draws on the operating mechanisms of language-computation models to propose hypotheses. A representative example of this type of work is that of Li et al. \cite{li2021cortical}. They used a cognitive model and a neural network model to study the neural mechanism of the brain when understanding pronouns. Different languages have different expressions of pronouns; for example, the pronunciation of Chinese pronouns is gender-neutral ("ta") while English pronouns are pronounced differently depending on gender ("she", "he", "it"). Thus, to explore whether the human brain adopts a general parsing strategy that is not affected by language, Li et al. collected fMRI data from Chinese and English native speakers while listening to the full text of "The Little Prince" in Chinese or English. A generalized linear model, commonly used in cognitive science, was used to calculate the brain regions related to pronoun resolution. Both Chinese and English listening materials significantly activated the left anterior middle temporal gyrus, left posterior middle temporal gyrus, and anterior and angular gyrus brain regions. 

To further explore the computing mechanism of the brain when parsing the relationship of reference, the researchers first constructed five computing models for pronoun reference resolution: the Hobbs model based on syntactic theory, the Centering model based on discourse theory, the ACT model based on memory theory, and the pronoun resolution model of ELMo and BERT based on neural networks. Next, they calculated the reference probabilities of each pronoun using the above models and correlated them with fMRI data. Only the ACT-R model based on memory theory could significantly predict neural activation data corresponding to the Chinese and English experimental materials, indicating that the brain adopts a language-independent general memory retrieval strategy when parsing the pronoun reference relationship. 

Another examples is Wehbe et al. \cite{wehbe2014aligning}, who proposed an analogy between the recurrent neural network language model (RNNLM) and the working mechanism of the reading brain. They found that the way the human brain works when reading a story is somewhat similar to how RNNLMs work when processing sentences. Additionally, Schrimpf et al. \cite{schrimpf2021neural} compared the association of 43 state-of-the-art neural network models with various neural activity datasets. They found that the model based on the language model and the transformer network structure can significantly predict the neural response, behavioral data, and neural response of the next word, indicating that the language system of the brain is optimized for predictive processing. See more recent work at survey \cite{abdou2022connecting}.

\subsection{Language computation methods inspired by language cognition} 
The deep-learning method based on neural networks has been highly praised in recent years. In a sense, it simulates the cognitive function of the biological brain. However, this method is not a mathematical model based on the working mechanism of the brain; thus, it is difficult to eliminate its dependence on largescale training samples. A large gap remains between language-computation models and human intelligence in terms of generalization and learning ability. The language-computation model inspired by the cognitive mechanism proposed in this article aims to study the language cognitive mechanism of the brain, analyze the relationship between the cognitive mechanism and machine language computation, and design a more intelligent language-computation model to complete various language-processing tasks.

Since the current research on the mechanism of language understanding in the brain is far less in-depth than other cognitive functions, most computational methods inspired by cognition are concentrated in the fields of visual cognition and machine learning, and less work has been done in the field of language. In this paper, existing cognitive-inspired language-computation methods are summarized into the following four categories:

\begin{enumerate}
    \item Cognitive function-inspired models
    
    To improve model performance when processing downstream tasks, we could borrow ideas of cognitive mechanisms such as brain representation, learning, attention, and memory to build new or improve existing computational models so that (part of) the model has a structure similar to the brain.
    
    For example, inspired by humans selectively looking at or skipping certain words when reading sentences, Wang et al. \cite{wang2017learning} proposed a sentence-representation model inspired by the human attention mechanism. This method utilizes the predictors of eye-movement signals (i.e., lexical surprisal and part-of-speech labels) to build attention modules. It introduces their results as weights into the sentence representation learning model. The results show that the attention module assigns higher attention weights to important words, and the weight results are significantly correlated with human reading time. In addition, the attention module can significantly improve the performance of sentence representation on several downstream tasks.

    In addition, Liu et al. \cite{liu2017attention} used the human attention mechanism to improve the performance of image description generation models. Sun et al. \cite{sun2018memory} proposed a small data word representation learning method based on memory enhancement. Finally, Han et al. \cite{han2020continual} proposed a continuous learning approach based on episodic memory activation and memory consolidation.

    \item Cognitive data-enhanced models
    
    We can use brain neural activity, neuroimaging, or behavioral data as an additional modality, which can provide different information than the existing data. Therefore, fusing these two data during model training could improve model performance.

    For example, Klerke et al. \cite{klerke2016improving} proposed a multitask learning approach to incorporate eye-tracking data into a sentence-compression task. They utilized a three-layer bidirectional recurrent network model with the bottom layer predicting eye-movement timing and the top layer predicting sentence compression. The results show that this multitask learning method can effectively introduce eye-movement data into the sentence-compression task and improve the performance of the model. Tiwalayo et al. \cite{eisape2020cloze} fused the probability distribution of the next word predicted by humans with that of the next word predicted by the language model, which effectively improved the performance of the language model.
    
    In addition, Malmaud et al. \cite{malmaud2020bridging} introduced predicted eye-movement time into reading-comprehension tasks. Barrett et al. \cite{barrett2016weakly} added eye-movement data as a feature to part-of-speech tagging and named entity recognition models. Mishra et al. \cite{mishra2017learning} applied eye-tracking data to improve the quality of sentiment-analysis models. Additionally, Fereidoni et al. \cite{fereidooni2020understanding} introduced fMRI data into vocabulary representation learning, and Roller et al. \cite{roller2013multimodal} and Wang et al. \cite{wang2018learning} introduced human behavior data (lexical association score) into multimodal vocabulary representation learning.

    \item Build models by simulating neurons
    
    Another way to build a more intelligent computing model is to simulate the structure and working mechanism of biological neurons or neural circuits from the underlying architecture of the model. For example, the fruit fly brain uses Kenyon cells to receive information from multiple sensory modalities. Specific neurons control the activation and inhibition states of these cells, so the fruit fly brain is a sparse high-dimensional representation of input information. 

Liang et al. \cite{liang2021can} formalized this information-encoding process and applied it to the task of word-representation learning. The experimental results indicate that the network can learn the static and context-dependent semantic representation of words, and its performance is comparable to other representation-learning methods. This method also represents the word as a sparse binary hash code, which requires fewer computational resources than other methods.
    
    In addition, this cognitively inspired method is commonly found in the research of general computing methods \cite{hassabis2017neuroscience}, such as memory networks \cite{weston2015memory}, neural Turing machines \cite{graves2014neural}, capsule networks \cite{sabour2017dynamic}, and plastic weight consolidation (Elastic Weight Consolidation) algorithms \cite{kirkpatrick2017overcoming}.

    \item Borrow methods from cognitive science to interpret models
    
    We could learn from or directly use research methods from cognitive science to interpret information encoded by neural network models.
    
    For example, Chien et al. \cite{chien2020mapping} used the timescale-mapping method commonly used in the field of neuroscience to study the information encoded by each neuron in the long short-term memory (LSTM) model. They inferred the neuron's function by observing the activation value of each neuron in the model for the following sentence when the normal sentence and the above fragment were randomly replaced. The logic behind it is that, if the function of a neuron encodes short-time-scale language information, then, when a small piece of text is replaced, its activation value change in the following sentence should be greater than that of neurons that encode long-term-scale language information. The study found that approximately 15\% of neurons are used to encode long-term scale information. Such neurons can be divided into two types: the controller responsible for connecting each neuron and the integrator (integrator) responsible for integrating long-distance information.

    In addition, inspired by neuroscience's approach to studying neuron encoding mechanisms, Lakretz et al. \cite{lakretz2019emergence} studied the working mechanism of each neuron in the LSTM model when completing tasks. Ivanva et al. \cite{ivanova2021probing} drew on the design methods of neuroscience probe tasks and proposed guidelines for designing probe tasks in machine-learning methods.
\end{enumerate}

\section{Discussion}
At this stage, researchers have achieved preliminary results in the two directions of using language-computation models to predict brain-activity data and inspiring language-computation models through the language understanding mechanism of the human brain. However, there is still a lack of granular and systematic research on the combination of language cognition and computation. For example, in terms of language comprehension in the human brain, it is not clear how people start from the most basic language unit and gradually build larger units until they finally understand the language. Moreover, there is still a lack of systematic and effective modeling methods to address this question. 

In terms of machine language understanding, the computation model has achieved super-human accuracy in many language-processing tasks. However, it is still far from human intelligence regarding common sense reasoning ability, autonomous learning ability, generalization ability, learning efficiency, interpretability, and reliability. There is no clear solution for how to learn from a deeper-brain language-understanding mechanism to build a more intelligent language-computation system.

\begin{figure}[htbp] \centering
    \includegraphics[width=0.8\textwidth]{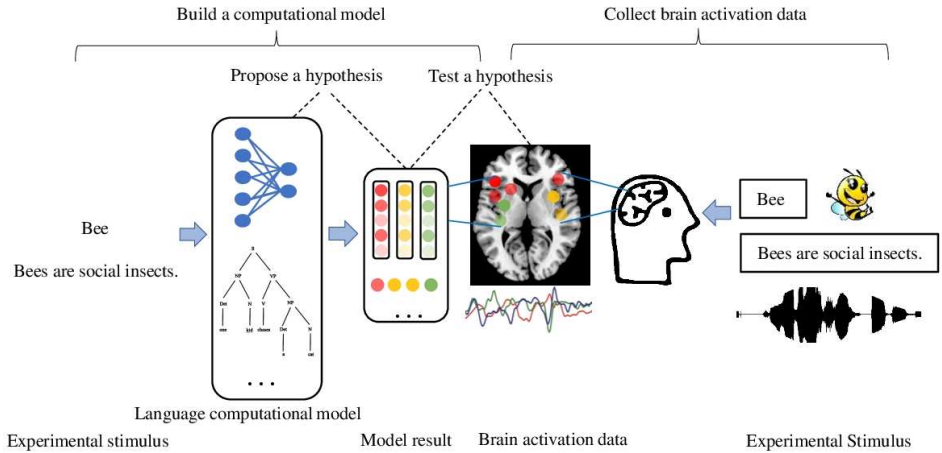}\\
    \caption{Schematic diagram of the cognitive experiment for natural language understanding driven by computational theory}\label{fig4}
    \end{figure}

In the future, the author believes that computational theory-driven language-comprehension cognitive experiments are promising for the study of human language comprehension, as shown in Figure \ref{fig4}.. In other words, research hypotheses are proposed based on the structure or results of computational models, and behavioral or brain activity data are verified. There are five important research directions.

\begin{enumerate}
    \item Collection of multilingual and multimodal neural activity data
    
    Most existing research on language cognition is limited to using a single data-collection method (such as fMRI or MEG) to study the specific language phenomenon of a certain language. This often leads to the problems of low robustness and poor repeatability of the conclusions drawn. Therefore, future language-cognition research should conduct verifications using multiple languages and multiple types of data \cite{wang2022fmri,wang2022synchronized}. Especially for studies combining computational models, the scale and quality of data directly determine the reliability of the results. Therefore, it is crucial to use both invasive and noninvasive tools to collect largescale high-quality neural activity data for different languages. At the same time, the opening and sharing of data is gradually becoming a trend, which will greatly promote the study of language cognition.
    
    \item Inspired new cognitive mechanism hypotheses
    
    The operation process of the language-computation model is transparent and global to a certain extent, and its calculation process is also visible. For instance, the vocabulary representation learned by the model, the calculation method of combining vocabulary representations into phrases and sentence representations, and the prediction and inference of certain calculation steps of the result are all observable. Explaining the working principle of the brain from the level of the computing mechanism is an important task of cognitive science. The author believes that, in the future, we can deeply explore whether the representation and computing modules in the computing model can indeed explain the neural activities of some brain regions in the process of language processing. If the neural activity of a brain region can be explained by a computational model, then the brain region can be considered to perform the computational functions clearly visible in the model. In other words, we can regard each module in different language computation models as a hypothesis of a brain computing mechanism and use cognitive science experiments to verify it.
    
    \item Correlating multiple linguistic variables and cognitive function
    
    The process of language comprehension is very complex, not only involving the processing of multiple language variables, such as morphology, syntax, and semantics, but also closely related to multiple cognitive functions, such as executive control, attention, and memory. Previous studies often eliminated the influence of other language variables and cognitive functions by strictly controlling experimental variables and only studied the effect of a certain language variable or cognitive function in an experiment. The author believes that language-cognition experiments combined with computational models can eliminate the research limitations above. For example, using computational models can separate different experimental variables and study the role of different language variables and cognitive functions based on neural activity data collected from natural texts\cite{wang2020probing,zhang2022probing,zhang2022does}. With the continuous improvement of the performance of language-computation methods based on neural network methods, it is increasingly accurate to use models to separate different language features so that the visual and auditory perception, multimodal information fusion, and language in different regions of the brain can be calculated on the same batch of data. Other functional mechanisms in understanding become possible.
    
    \item Analyzing the underlying computing mechanism of brain language understanding
    
    Most of the existing research on language cognition is based on linguistic theory, but there is a large gap between linguistics and neuroscience research. For example, regarding how the brain manipulates the most basic language units, linguistics mainly studies phrase structure and semantic combination while neuroscience focuses on neural oscillations and synchrony. This has led to the lack of a neural basis in the current research on language cognition, which cannot match the conclusions of neuroscience findings. With the continuous development of spike neural networks (spike neural networks) and oscillating neural networks (oscillatory neural networks), future computing models must be able to integrate the conclusions of neuroscience to simulate the working mode of underlying neurons. Manipulating language units to complete the task of language understanding provides a new solution for research linking linguistics and neuroscience.

    \item Exploring the mechanism of language learning and evolution
    
    As early as the 1980s, cognitive scientists used the connectionist model to explore what kind of model and what kind of data can simulate the human language-acquisition process. However, the computing power of the connectionist model at that time was quite limited, and it could only solve some case-specific and simple language tasks. Today, the language-processing ability of deep neural networks has made a qualitative leap compared to the 1980s, and there are more corpus records in the process of infant language acquisition. Therefore, it is possible to try to use computational models to explore the mechanism of language learning and evolution. It is even possible to explore in depth whether the cyclic connection, convolution operation, dot product attention mechanism, backpropagation algorithm, and so on in the language-computation model are also necessary links in the human brain's language understanding and computing process.
\end{enumerate}

\begin{figure}[htbp] \centering
    \includegraphics[width=0.8\textwidth]{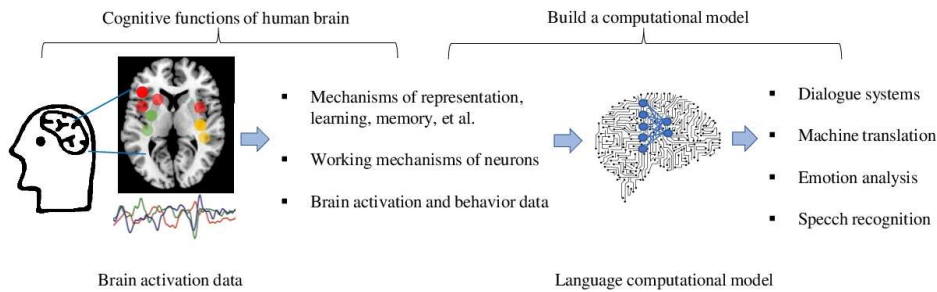}\\
    \caption{Schematic diagram of a language computational model inspired by the cognitive function of the human brain}\label{fig5}
    \end{figure}

With regard to machine language comprehension, as shown in Figure \ref{fig5}, research on language cognition suggests that the human brain may have many efficient processing methods. Thus, it has great potential to inspire the construction of a new generation of language-computation models regarding the cognitive mechanisms of representation, learning, and memory and the working mechanisms of neurons. The author believes that the following five aspects will become important development directions in the future.

\begin{enumerate}
    \item Representation and combination of text
    
    When encoding the meaning of concepts, the brain uses different representations for different types. For example, when reading nouns and verbs, the brain activates different brain networks, showing the characteristics of distributed coding. When observing a specific or familiar person, a specific neuron in the brain is activated. When encoding the syntactic structure, the brain will use different combination methods for different types of phrases, use hierarchical encoding methods such as tree structures to guide the combination sequence of words, and use parallel processing to encode multiple levels of language unit (words, phrases, sentences, etc.) information. With this encoding method, the human brain stores and calculates the meaning of the text very efficiently. It is also closely related to the ability of humans to "infer other cases from one instance" and learn quickly. Future language computation models can learn from this mechanism; combine symbolic and distributed representation methods; and adopt a combination of diversity, hierarchy, and parallelism to learn text representation and combination models.
    
    \item Continuous language learning
    
    Humans have the ability of continuous and small-sample learning in childhood. For example, if a 2-year-old child is shown a picture of a giraffe, the child can recognize other pictures of giraffes. This ability can be transferred to tasks such as recognizing pictures of other animals and finding the text corresponding to the picture. This learning ability is closely related to the human memory system, and the result of learning new information is memory. Different types of information are processed and stored by different memory systems, and information such as sounds and images are stored by sensory memory and maintained in a short time. The complex and structured memory system of the human brain ensures the efficient organization of massive data and rapid extraction when necessary. All these mechanisms could be learned by computation models to improve performance.

    \item Interactive learning of language
    
    The best existing general-purpose language-computation models use predicting the next word as an objective function. They are trained in massive texts and achieve excellent performance in multiple tasks. The difference is that humans often learn language by interacting with others, which is a more effective way to learn and improve language ability than analyzing and memorizing language structures. Drawing on this interactive learning method, in addition to using text information as a supervisory signal, future language-computation models can also obtain feedback from the structure or output results of other models and learn in continuous interaction with each other.

    \item Multimodal information fusion
    
    Closely related to interactive learning is the comprehensive processing of multiple-modality information. The human language-learning environment is a multimodal system. Humans are better at processing multimodal than single-modal information, and the processing speed is faster for multimodal information than for single-modal information \cite{holler2019multimodal}. Therefore, the author believes that, in the fusion of multimodal information, brain-inspired computing models are an important research direction in the future. For example, according to the "Hub and Spoke" theory \cite{patterson2016hub}, concepts are represented by multimodal information such as vision, hearing, smell, and somatosensory information, and there is a semantic center that encodes modal-independent information to associate different modalities of information. The information between different modalities is complementary and mutually verifiable and, when combined, can represent more abundant information. The fusion mechanism of multimodal information can also learn from the abovementioned "central" mechanism and design a modality-independent module to integrate and correlate different types of information.

    \item Interpretability of computation models
    
    Cognitive science designs experiments to analyze the working mechanism of the human brain. This research method and cognitive experimental data can also be used to analyze or evaluate the working mechanism of language-computation models and inspire new model interpretability methods. For example, referring to the comparative analysis method often used in language-cognition experiments, two groups of experimental materials are designed so that they differ only in a certain language attribute. For example, groups of sentences with high and low syntactic complexity that are basically the same in terms of sentence length, sentence meaning, and so on are input into the calculation model, and we observe whether the effect of the model is consistent with the syntactic complexity degree correlation. If the activation of some nodes in the network is significantly stronger when encoding high- than low-complexity sentences, then these neurons are responsible for encoding syntactic information; otherwise, they are not responsible for encoding syntactic information.
\end{enumerate}

\section{Conclusion} 
Language cognition is one of the core issues in cognitive and brain sciences. It is of great significance not only to reveal the basis of  language intelligence and the working mechanisms of the brain but also to help promote the development of brain-inspired language intelligence technology. Simultaneously, new ideas and technologies in language-computation research can also provide important references and support for the study of language cognition. Therefore, future research on language cognition and computation will inevitably produce a closer combination, and the prospects of interdisciplinary research in this area are worth exploring.


\end{document}